\documentclass[10pt,journal,compsoc]{IEEEtran}

\usepackage{times}
\usepackage{graphicx} 

\usepackage{algorithm}
\usepackage{algorithmic}

\usepackage{hyperref}

\usepackage{times}
\usepackage{epsfig}
\usepackage{graphicx}
\usepackage{amsmath}
\usepackage{amssymb}
\usepackage{booktabs}
\usepackage{etex}
\usepackage{footnote}
\usepackage{tablefootnote}
\usepackage{enumerate}
\usepackage{enumitem}
\usepackage{algorithmic}

\usepackage{graphicx}
\usepackage{latexsym}
\usepackage{amsmath}
\usepackage{amssymb}
\usepackage{subcaption}
\usepackage{multirow}

\usepackage{color}
\usepackage[dvipsnames]{xcolor}

\usepackage{epstopdf}
\usepackage{array}
\newcolumntype{C}[1]{>{\centering\arraybackslash}m{#1}}

\newcommand{\gr}{{\rm grad}}

\usepackage{url}

\usepackage{amssymb}
\usepackage{amsfonts}
\usepackage{amsthm}
\usepackage{MnSymbol}

\usepackage[cmtip,all]{xy}
\usepackage[all,cmtip]{xy}

\newcommand{\Mi}{\mathcal{M}_{\iota}}
\newcommand{\Ci}{\mathcal{C}_{\iota}}
\newcommand{\Oi}{\omega_{\iota}}

\newtheorem{theorem}{Theorem}[section]
\newtheorem{lemma}[theorem]{Lemma}

\newtheorem{corr}[theorem]{Corollary}

\theoremstyle{definition}
\newtheorem{definition}[theorem]{Definition}
\theoremstyle{definition}

\newtheorem{example}[theorem]{Example}

\theoremstyle{remark}

\theoremstyle{remark}

\theoremstyle{remark}

\newcommand*{\QEDbs}{\hfill\ensuremath{\blacksquare}}%
\begin{document}

\title{Optimization on Product Submanifolds of Convolution Kernels}

\author{Mete~Ozay,~Takayuki~Okatani\\
	Graduate School of Information Sciences, Tohoku University, Sendai, Miyagi, Japan.
	{\tt\small \{mozay,okatani\}@vision.is.tohoku.ac.jp} }


\maketitle

\begin{abstract} 
Recent advances in optimization methods used for training convolutional neural networks (CNNs) with kernels, which are normalized according to particular constraints, have shown remarkable success. This work introduces an approach for training CNNs using ensembles of joint spaces of kernels constructed using different constraints. For this purpose, we address a problem of optimization on ensembles of products of  submanifolds (PEMs) of convolution kernels. To this end, we first propose three strategies to construct ensembles of PEMs in CNNs. Next, we expound their geometric properties (metric and curvature properties) in CNNs. We make use of our theoretical results by developing a geometry-aware SGD algorithm (G-SGD) for optimization on ensembles of PEMs to train CNNs. Moreover, we analyze convergence properties of G-SGD considering geometric properties of PEMs. In the experimental analyses, we employ G-SGD to train CNNs on Cifar-10, Cifar-100 and Imagenet datasets. The results show that geometric adaptive step size computation methods of G-SGD can improve training loss and  convergence properties of CNNs. Moreover, we observe that classification performance of baseline CNNs can be boosted using G-SGD on ensembles of PEMs  identified by multiple constraints. 
\end{abstract} 

\section{Introduction}
\label{intro}

In the recent works  \cite{unit_evol,norm_prop,parseval,Henaff,Huang_2017_ICCV,icml2015_huanga15,SNN,IGAN,w_norm,AAAI1714830}, several methods have been suggested to train deep neural networks using kernels (weights) with various normalization constraints to boost their performance. Spaces of normalized kernels have been explored using Riemannian manifolds (e.g. the Stiefel), and stochastic optimization algorithms have been employed to train CNNs using kernel manifolds in \cite{RBN,haaai,huang162,oo16}. 



In this work, we suggest an approach for training CNNs using multiple constraints on kernels in order to learn a richer set of features compared to the features learned using single constraints. We address this problem by optimization on ensembles of products of different kernel submanifolds (PEMs) that are identified by different constraints of kernels. However, if we employ the aforementioned Riemannian SGD algorithms \cite{sgdman,RBN,oo16} on PEMs to train CNNs, then we observe early divergence, vanishing and exploding gradients problems. Therefore, we elucidate geometric properties of PEMs to assure convergence to local minima while training CNNs using our proposed geometry-aware stochastic gradient descent (G-SGD). Our contributions are summarized as follows:
\begin{enumerate}

\item We explicate the geometry of space of convolution kernels defined by multiple constraints. For this purpose, we explore the relationship between geometric properties of PEMs, such as sectional curvature, geodesic distance, and gradients computed at PEMs, and those of component submanifolds of convolution kernels in CNNs (see Lemma~\ref{lemma32} in Section~\ref{sec3}).  

\item We propose an SGD algorithm (G-SGD) for optimization on different ensembles of PEMs (Section~\ref{sec3}) by generalizing the SGD methods employed on kernel submanifolds \cite{haaai,huang162,oo16}. Next, we explore the effect of geometric properties of the PEMs on the convergence of the G-SGD using our theoretical results. Then, we employ the results for adaptive computation of step size of the SGD (see Theorem~\ref{thm33} and Corollary~\ref{corr34}). Moreover, we provide an example for computation of a step size function for optimization on PEMs identified by the sphere (Corollary~\ref{corr34}). In addition, we propose three strategies in order to construct ensembles of identical and non-identical kernel spaces according to their employment on input and output channels in CNNs in Section~\ref{sec2}. To the best of our knowledge, our proposed G-SGD is the first algorithm which performs optimization on different ensembles of PEMs to train CNNs with convergence properties.


\item We experimentally analyze convergence properties and classification performance of CNNs on benchmark image classification datasets such as Cifar 10/100 and Imagenet, using various manifold ensemble schemes (Section~\ref{sec:exp}). In the results, we observe that G-SGD employed on ensembles of PEMs can boost baseline state-of-the-art performance of CNNs.  
\end{enumerate}


Proofs of the theorems, additional results, and implementation details of the algorithms and datasets are given in the supplemental material.

\section{Construction of Ensembles of PEMs}
\label{sec2}

Suppose that we are given a set of training samples ${S=\{s_i= (\mathbf{I}_i,y_i) \}_{i=1}^N}$ of a random variable $s$ drawn from a distribution $\mathcal{P}$ on a measurable space $\mathfrak{S}$, where $y_i $ is a class label of the $i^{th}$ image $\mathbf{I}_i$. An $L$-layer CNN consists of a set of tensors $\mathcal{W} = \{\mathcal{W}_l \}_{l=1}^L$, where ${\mathcal{W}_l = \{ \mathbf{W}_{d,l} \in \mathbb{R}^{A_l \times B_l \times C_l} \} _{d=1} ^{D_l}}$, and ${\mathbf{W}_{d,l} = [W_{c,d,l} \in \mathbb{R}^{A_l \times B_l}]_{c=1}^{C_l}}$  is a tensor\footnote{We use shorthand notation for matrix concatenation such that $[W_{c,d,l}  ]_{c=1}^{C_l} \triangleq [W_{1,d,l}, W_{2,d,l}, \cdots,W_{C_l,d,l}]$.} composed of kernels (weight matrices) $W_{c,d,l} $ constructed at each layer ${l=1,2,\ldots,L}$, for each $c^{th}$ channel $c=1,2,\ldots,C_l$ and each $d^{th}$ kernel $d=1,2,\ldots,D_l$. At each $l^{th}$ convolution layer, we compute a feature representation $f_l(\mathbf{X}_l;\mathcal{W}_l)$ by compositionally employing non-linear functions, and convolving an image $\mathbf{I}$ with kernels by  
\begin{equation}
f_l(\mathbf{X}_l;\mathcal{W}_{l}) = f_l(\cdot;\mathcal{W}_l) \circ  \cdots \circ f_1(\mathbf{X}_1;\mathcal{W}_{1}),
\label{eq:comp_rep}
\end{equation}
where ${\mathbf{X}_1 := \mathbf{I}}$ is an image for ${l=1}$, and $\mathbf{X}_{l} = [ X_{c,l}]_{c=1}^{C_l}$. The $c^{th}$ channel of the data matrix $X_{c,l}$ is convolved with the kernel ${W}_{c,d,l}$ to obtain the $d^{th}$ feature map ${ X_{c,l+1} : = \hat{X}_{d,l}}$ by $\hat{X}_{d,l} = {W}_{c,d,l} \ast X_{c,l}, \forall c, d, l$ \footnote{We ignore the bias terms in the notation for simplicity.}. Given a batch of samples $\mathfrak{s} \subseteq S$, we denote a value of a classification loss function for a kernel $\omega \triangleq W_{c,d,l}$ by $\mathcal{L}(\omega,\mathfrak{s})$, and the loss function of kernels $\mathcal{W}$ utilized in the CNN by $\mathcal{L}(\mathcal{W},\mathfrak{s})$. Assuming that  $\mathfrak{s}$ contains a single sample, an expected loss or cost function of the CNN  is computed by
\begin{equation}
\mathcal{L}(\mathcal{W}) \triangleq E_{\mathcal{P}} \{ {\mathcal{L}}(\mathcal{W},s) \} = \int {\mathcal{L}}(\mathcal{W},s) d \mathcal{P}.
\label{eq:expected_cost}
\end{equation}
The expected loss $\mathcal{L}(\omega)$ for $\omega$ is computed by 
\begin{equation}
\mathcal{L}(\mathcal{\omega}) \triangleq E_{\mathcal{P}} \{ {\mathcal{L}}(\mathcal{\omega},s) \} = \int {\mathcal{L}}(\mathcal{\omega},s) d \mathcal{P}.
\label{eq:expected_w_scost}
\end{equation}
 For a finite set of samples $S$,  $\mathcal{L}(\mathcal{W})$ is approximated by an empirical loss $\frac{1}{|S|} \sum_{i=1}^{|S|} \mathcal{L}(\mathcal{W},s_i)$, where $|S|$ is the size of $S$ (similarly, $\mathcal{L}(\omega)$ is approximated by the empirical loss for $\omega$). Then, feature representations are learned by solving
\begin{equation}
\min_{\mathcal{W}} \mathcal{L}(\mathcal{W})
\label{eq:opt1}
\end{equation}
using an SGD algorithm. In the SGD algorithms employed on kernel submanifolds \cite{haaai,huang162,oo16}, each kernel is assumed to reside on an embedded kernel submanifold $\mathcal{M}_{c,d,l}$ at the $l^{th}$ layer of a CNN, such that ${\omega \in \mathcal{M}_{c,d,l}}, \forall c,d$. In this work, we propose a geometry-aware SGD algorithm (G-SGD), by generalizing the SGD algorithms \cite{haaai,huang162,oo16} for optimization on ensembles of different products of the kernel submanifolds, which are defined next.

\begin{definition}[Products of embedded kernel submanifolds of convolution kernels (PEMs) and their ensemble] 
	Suppose that ${\mathcal{G}_l = \{ \mathcal{M}_{\iota}: \iota \in \mathcal{I}_{\mathcal{G}_l} \}}$ is an ensemble of Riemannian kernel submanifolds $\mathcal{M}_{\iota}$ of dimension $n_{\iota}$, which is identified by a set of indices $\mathcal{I}_{\mathcal{G}_l}, \forall {l=1,2,\ldots,L}$. More concretely, $\mathcal{I}_{\mathcal{G}_l}$ contains indices each of which represents an identity number ($\iota$) of a kernel that resides on a manifold $\mathcal{M}_{\iota}$ at the $l^{th}$ layer. In addition, a subset ${\mathcal{I}_{{G}_l}^m \subseteq \mathcal{I}_{\mathcal{G}_l}}, {m =1,2,\ldots,M}$, is used to determine a subset ${G}^m_l \subseteq \mathcal{G}_l$ of kernel submanifolds  which will be aggregated to construct a PEM, and satisfies the following properties:
	\begin{itemize}
	\item Each subset of indices contains at least one kernel such that ${\mathcal{I}_{{G}_l}^m} \neq \emptyset$, for each $m=1,2,\ldots,M$.	
	\item The set of indices $\mathcal{I}_{\mathcal{G}_l}$ is covered by the subsets ${\mathcal{I}_{{G}_l}^m}$ such that $\mathcal{I}_{\mathcal{G}_l} = \bigcup \limits _{m=1} ^M {\mathcal{I}_{{G}_l}^m}$. 
	\item If kernels are not shared among PEMs such that ensembles are constructed using non-overlapping sets, then $\mathcal{I}_{G_l}^m \cap \mathcal{I}_{{G}_l}^{\bar{m}} = \emptyset$ for $m \neq \bar{m}$.
	\item If kernels are shared among PEMs such that ensembles are constructed using overlapping sets, then ${\mathcal{I}_{G_l}^m \cap \mathcal{I}_{{G}_l}^{\bar{m}} \neq \emptyset}$ for $m \neq \bar{m}$.
	\end{itemize} 
	A $G^m_l$ product manifold of convolution kernels ($G^m_l$-PEM) constructed at the $l^{th}$ layer of an $L$-layer CNN, denoted by $\mathbb{M}_{G^m_l}$, is a product of embedded kernel submanifolds belonging to ${G}^m_l$ which is computed by
\begin{equation}
\mathbb{M}_{G^m_l} = \bigtimes_{\iota \in \mathcal{I}^m_{{G}_l}} \mathcal{M}_{\iota} ,
\label{eq:prod_man}
\end{equation}
where $\bigtimes$ is the topological Cartesian product, and therefore $\mathbb{M}_{G^m_l}$ is a product topology. Each ${\mathcal{M}_{\iota} \in  {G}^m_l}$ is called a component submanifold of $\mathbb{M}_{G^m_l}$. A kernel $\omega_{G^m_l} \in \mathbb{M}_{G^m_l}$ is then obtained by concatenating kernels belonging to $\mathcal{M}_{\iota}$, $\forall \iota \in \mathcal{I}^m_{{G}_l}$, using  ${\omega_{G^m_l} = (\omega_1, \omega_2, \cdots, \omega_{|\mathcal{I}^m_{{G}_l}|})}$, where $|\mathcal{I}^m_{{G}_l}|$ is the cardinality of $\mathcal{I}^m_{{G}_l}$. A $\mathcal{G}_l$-PEM is called an ensemble of PEMs constructed using \eqref{eq:prod_man} for $m=1,2,\ldots,M$. 
\QEDbs
\end{definition}

\begin{figure*}[t!]
	\centering
	\includegraphics[width=6.850in]{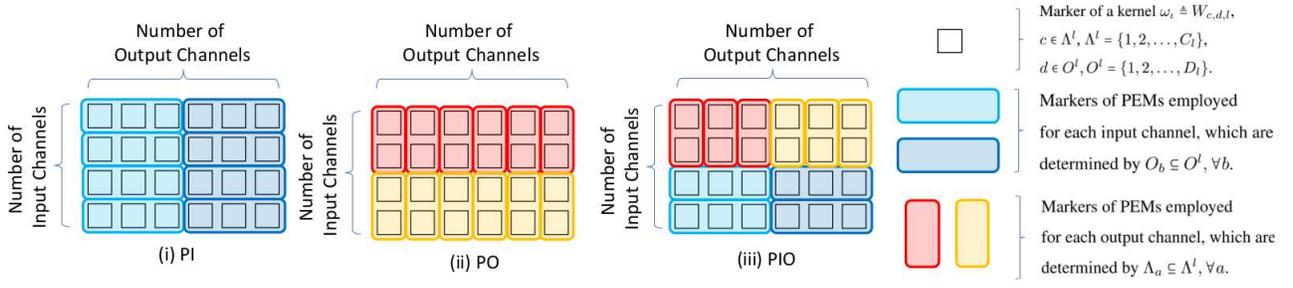}%
	\caption{An illustration for employment of the proposed PI, PO and PIO strategies at the $l^{th}$ layer of a CNN.}
	\label{fig_block}
\end{figure*}

We compute a PEM $\mathbb{M}_{G^m_l}$ using component submanifolds $\mathcal{M}_{\iota}$ in \eqref{eq:prod_man} utilizing ${\mathcal{I}_{{G}_l}^m \subseteq \mathcal{I}_{\mathcal{G}_l}}, m=1,2,\ldots,M$, and construct ensembles of PEMs $\mathcal{G}_l$ using $\mathcal{I}_{\mathcal{G}_l}$. Recall that, at each $l^{th}$ layer of an {$L$-layer} CNN, we compute a convolution kernel ${\omega_{\iota} \triangleq W_{c,d,l}}$, ${c \in \Lambda^l},  \Lambda^l=\{1,2,\ldots,C_l\}$, ${d \in O^l}$, $O^l=\{1,2,\ldots,D_l \}$. We first choose $\mathfrak{A}$ subsets of indices of input channels ${\Lambda_a \subseteq \Lambda^l}, a=1,2,\ldots,\mathfrak{A}$ and $\mathfrak{B}$ subsets of indices of output channels $O_b \subseteq O^l, b=1,2,\ldots,\mathfrak{B}$, such that $\Lambda^l = \bigcup \limits _{a=1} ^\mathfrak{A} \Lambda_a$ and $O^l = \bigcup \limits _{b=1} ^\mathfrak{B} O_b$. Then, we propose three strategies for determination of index sets (see Figure~\ref{fig_block});
\begin{enumerate}[leftmargin=*] 
	\item PEMs for input channels (PI): For each $c^{th}$ input channel, we construct $\mathcal{I}_{\mathcal{G}_l} = \bigcup \limits _{c=1} ^{C_l} \mathcal{I}_{{G}_l} ^c   $, where  ${\mathcal{I}_{{G}_l} ^c =  O_b \times \{c\} }$ and the Cartesian product ${O_b \times \{c\}} $ preserves the input channel index, $\forall b,c$. 	
	\item PEMs for output channels (PO): For each $d^{th}$ output channel, we construct $\mathcal{I}_{\mathcal{G}_l} = \bigcup \limits _{d=1} ^{D_l} \mathcal{I}_{{G}_l} ^d  $, where   ${\mathcal{I}_{{G}_l} ^d = \Lambda_a \times \{d\}  }$ and the Cartesian product $\Lambda_a \times \{d\} $ preserves the output channel index, $\forall a,d$. 	
	\item PEMs for input and output channels (PIO):  We construct $\mathcal{I}_{{G}_l}^{a,b} =  \mathcal{I}_{{G}_l}^{a} \cup  \mathcal{I}_{{G}_l}^{b}$, where $ \mathcal{I}_{{G}_l}^{a} = \{ \Lambda_a \times a \}$ and $ {\mathcal{I}_{{G}_l}^{b} = \{ O_b \times b\} }$  such that $\mathcal{I}_{\mathcal{G}_l} = \bigcup \limits _{a=1, b=1} ^{\mathfrak{A},\mathfrak{B}} \mathcal{I}_{{G}_l} ^{a,b}$.	
\end{enumerate}

\begin{example}
An illustration of employment of PI, PO and PIO at the $l^{th}$ layer of a CNN is given in Figure~\ref{fig_block}. Suppose that we have a kernel tensor of size $3 \times 3 \times 4 \times 6$ where the number of input and output channels is $4$ and $6$. In total, we have ${4*6=24}$ kernel matrices of size $3 \times 3$. An example of construction of an ensemble of PEMs is as follows. 
\begin{enumerate}[leftmargin=*] 
\item PI: For each of 4 input channels, we split a set of 6 kernels associated with 6 output channels into two subsets of 3 kernels. Choosing the sphere (Sp) for the first subset, we construct a PEM  as a product of 3 Sp using \eqref{eq:prod_man}. That is, each of 3 component manifolds ${\mathcal{M}_{\iota}}, {\iota  = 1,2,3}$, of the PEM is a sphere. Similarly, choosing the Stiefel (St) for the second subset, we construct another PEM as a product of 3 St (each of 3 component manifolds ${\mathcal{M}_{\iota}}, \iota  = 1,2,3$, of the second PEM is a Stiefel manifold.). Thus, at this layer, we construct an ensemble of 4 PEMs of 3 St and 4 PEMs of 3 Sp.
\item PO: For each of 6 output channels, we split a set of 4 kernels corresponding to the input channels into two subsets of 2 kernels. We choose the Sp for the first subset, and we construct a PEM as a product of 2 Sp using \eqref{eq:prod_man}. We choose the St for the second subset, and we construct a PEM as a product of 2 St. Thereby, we have an ensemble consisting of 6 PEMs of St and 6 PEMs of Sp.
\item PIO: We split the set of 24 kernels into 10 subsets. For each of 6 output channels, we split the set of kernels corresponding to the input channels into 3 subsets. We choose the Sp for 2 subsets each containing 3 kernels, and 3 subsets each containing 2 kernels. We choose the St similarly for the remaining subsets. Then, our ensemble contains 5 PEMs of St and 5 PEMs of Sp.
\end{enumerate}
\end{example}
Our framework can be used to model both overlapping and non-overlapping sets. If ensembles are constructed using overlapping sets, then kernels having different constraints can be applied to the same input or output channels. For example, kernels belonging to a PEM of 3 St and kernels belonging to a PEM of 3 Sp can be applied to the same output (input) channel for PI (PO) in the previous example (see Figure~\ref{fig_block}). More complicated configurations can be obtained using PIO. In the experiments, we selected non-overlapping sets for simplicity. We consider theoretical and experimental analyses of overlapping sets as a future work.

\section{Optimization on Ensembles of PEMs using Geometry-aware SGD in CNNs}
\label{sec3}
If an SGD is employed on non-linear kernel submanifolds, then the gradient descent is generally performed by three steps;  i) projection of gradients on tangent spaces of the submanifolds, ii) movement of kernels on the tangent spaces in the gradient descent direction, and iii)  projection of the moved kernels onto the submanifolds \cite{oo16}. These steps are determined according to the geometric properties of the submanifolds, such as sectional curvature and metric properties. For example, the Euclidean space has zero sectional curvature, i.e. it is not curved (\textit{flat}). Thereby, these steps can be performed using a single step if an SGD employs kernels residing on the Euclidean space. However, if kernels belong to the unit sphere, then the kernel space is curved by constant positive curvature. Moreover, a different tangent space is computed at each kernel located on the sphere. Therefore, nonlinearity of operations and transformations applied on kernels implied by curvature and metric of kernel spaces are used for gradient descent in the aforementioned three steps. In addition, martingale properties of stochastic processes defined by kernels are determined by geodesics, metrics, gradients projected at tangent spaces and injectivity radius of kernel spaces (see proofs of Theorem~\ref{thm33} and Corollary~\ref{corr34} in the supp. mat. for details). 

\begin{figure}[t!]
	\centering
	\begin{subfigure}[b]{0.15\textwidth}		
		\label{fig:sphere}%
		\includegraphics[width=1.3in]{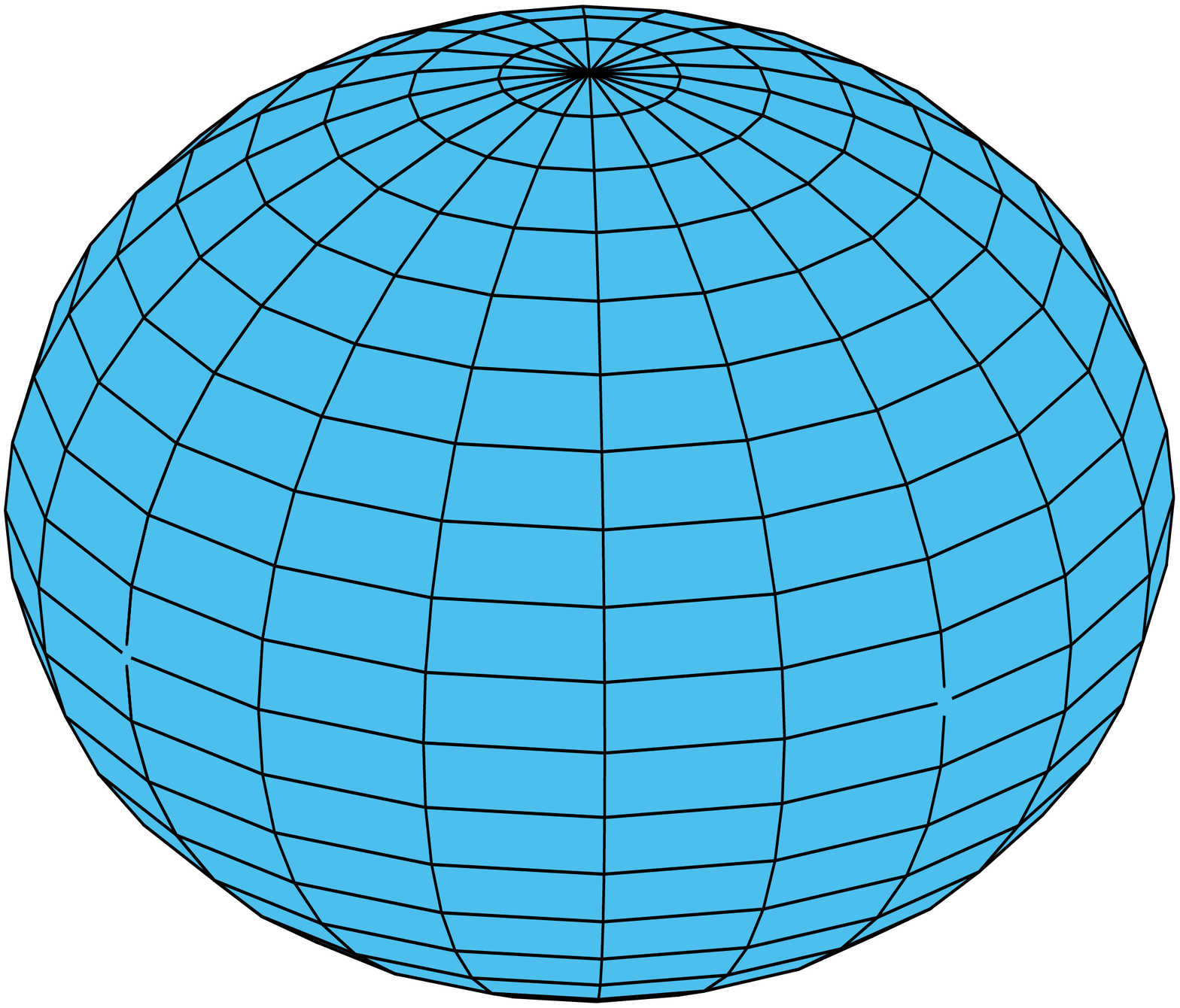}%
		\caption{$\mathbb{S}^2$.}
	\end{subfigure}
	\begin{subfigure}[b]{0.15\textwidth}		
		\label{fig:torus}%
		\includegraphics[width=1.0in]{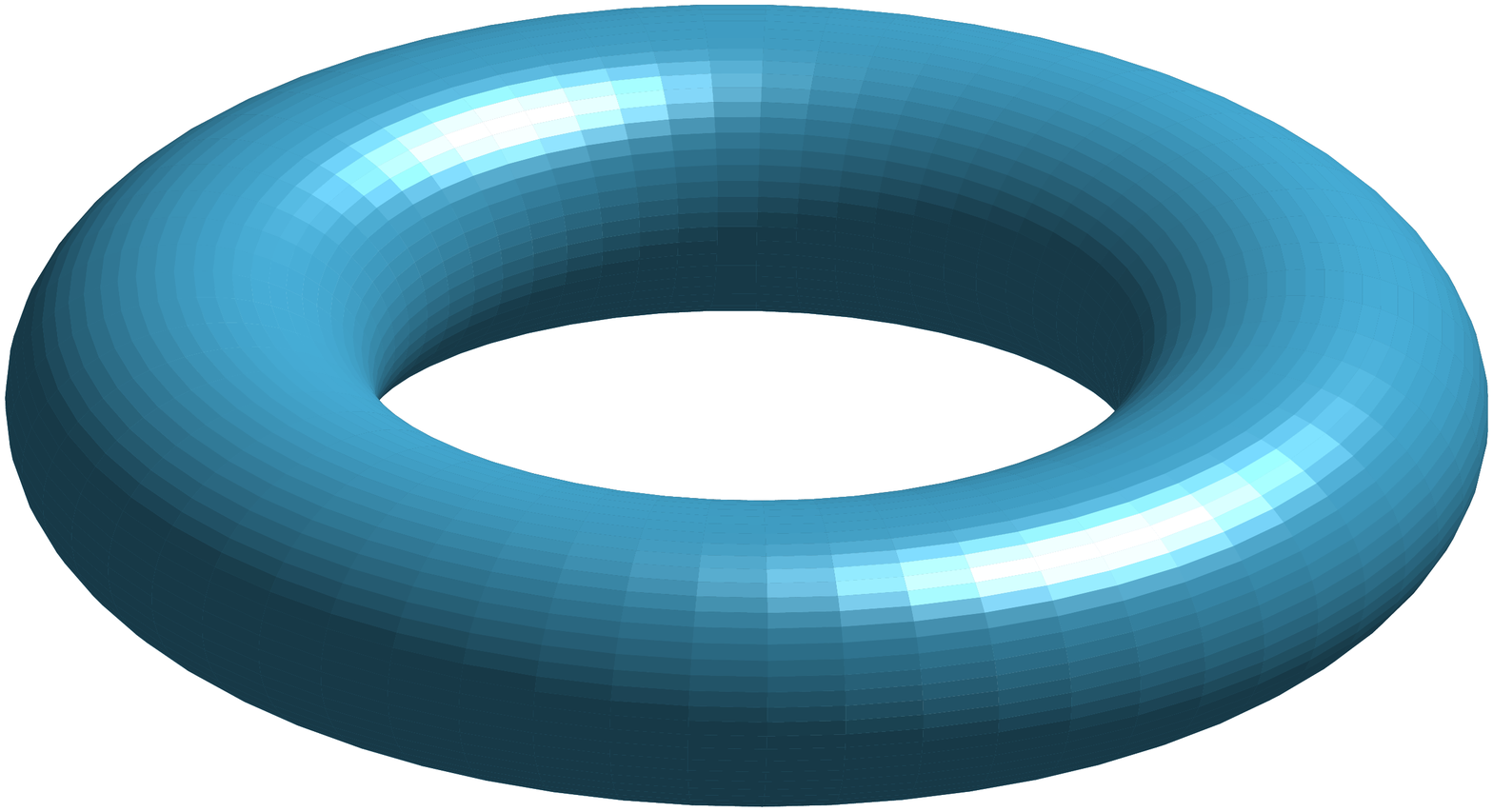}%
		\caption{$\mathbb{T}^2 = \mathbb{S}^1 \times \mathbb{S}^1$.}
	\end{subfigure}
	\begin{subfigure}[b]{0.15\textwidth}
		\label{fig:cylinder}%
		\includegraphics[width=1.3in]{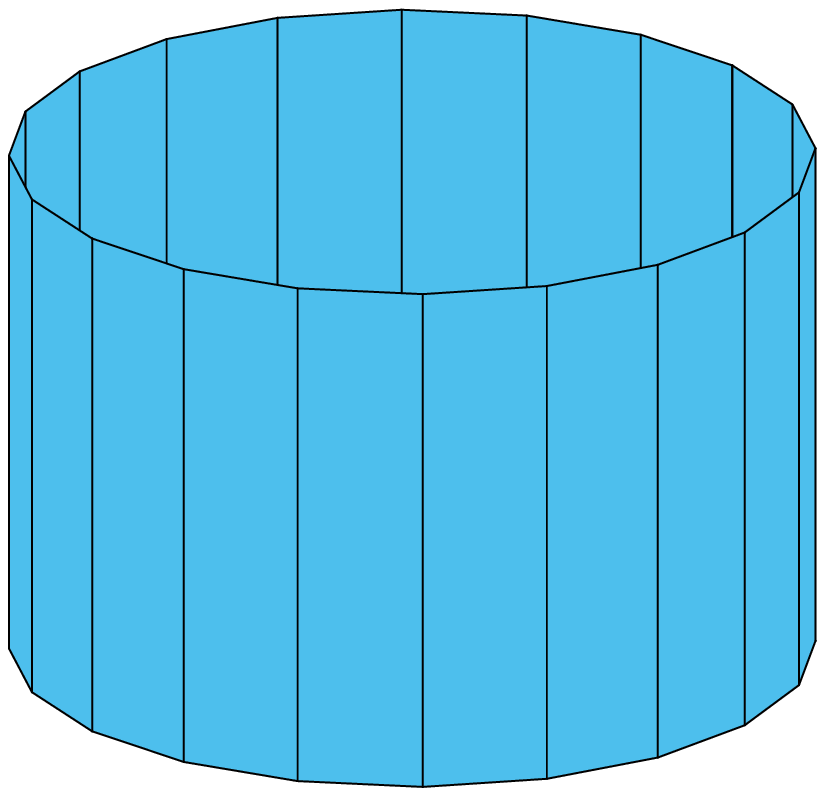}%
		\caption{$\mathbb{S}^1 \times \mathbb{R}$.}
	\end{subfigure}
	\caption{(a) An orthonormalized convolution kernel ${\omega \in \mathbb{R}^{3 \times 1} }$ (${\omega \in \mathbb{R}^{A \times B}}$) resides on a two-sphere $\mathbb{S}^2$ ($\mathbb{S}^{AB-1}$) which has constant positive sectional curvature, $1$. (b) A kernel $\omega = (\omega_1, \omega_2)$, where each $\omega_{\iota} \in \mathbb{R}^{2 \times 1}, \iota = 1,2$, belongs to a circle $\mathbb{S}^1$, resides on a two-torus $\mathbb{T}^2$ with varying curvature. (c) If $\omega_{1} \in \mathbb{S}^{1}$ ($\omega_{1} \in \mathbb{S}^{p}$) and $\omega_{2} \in \mathbb{R}$ ($\omega_{2} \in \mathbb{R}^{q-p}$), then $\omega$ resides on a cylinder $\mathbb{S}^1 \times \mathbb{R}$ with varying curvature ($q$-cylinder $\mathbb{S}^p \times \mathbb{R}^{q-p}$). In general, if a convolution kernel $\omega \in \mathbb{M}$ resides on a PEM $\mathbb{M}$, then $\mathbb{M}$ does not admit a metric with negative sectional curvature (see Lemma~\ref{lemma32}). Therefore, geometric properties of PEMs, which can be composed of non-identical component submanifolds, may crucially affect convergence of SGD methods for training of CNNs  (see Theorem~\ref{thm33} and Corollary~\ref{corr34}). }
	\label{fig1}
\end{figure}

Geometric properties of PEMs can be different from that of the component submanifolds of PEMs, even if they are constructed using identical submanifolds. For example, we observe locally varying curvatures when we construct PEMs of spheres (see Figure~\ref{fig1}). Kernel spaces with more complicated geometric properties can be obtained using the proposed strategies (PI, PO, PIO), especially by constructing ensembles of PEMs of non-identical submanifolds (see Section~\ref{sec:exp} for details and examples). Thus, as the complexity of geometry of kernel spaces increases, their effect on performance and convergence of SGD gradually increases. 

In order to address these problems and consider geometric properties of kernel submanifolds for training of CNNs, we propose a geometry aware SGD (G-SGD). We employ metric properties of PEMs to perform gradient descent steps of G-SGD, and use curvature properties PEMs to explore convergence properties of G-SGD. We  explore metric and curvature properties of PEMs in the next theorem. 




\begin{definition}[Sectional curvature of component submanifolds]
	Let $\mathfrak{X}(\Mi)$ denote the set of smooth vector fields on $\Mi$. The sectional curvature of $\Mi$ associated with a two dimensional subspace $\mathfrak{T} \subset \mathcal{T}_{\Oi}\Mi$  is defined by 
	\begin{equation}
	\mathfrak{c}_{\iota} = \frac{\left\langle \Ci(X_{\Oi},Y_{\Oi})Y_{\Oi},X_{\Oi}  \right\rangle}{\left\langle  X_{\Oi} , X_{\Oi} \right\rangle \left\langle  Y_{\Oi} , Y_{\Oi} \right\rangle  -   \left\langle X_{\Oi} ,Y_{\Oi} \right\rangle^2}
	\end{equation}
	where $\Ci(X_{\Oi},Y_{\Oi})Y_{\Oi}$ is the Riemannian curvature tensor\footnote{Additional definitions are given in the supp. mat.}, $\left\langle \cdot,\cdot \right\rangle$ is an inner product, ${X_{\Oi} \in \mathfrak{X}(\Mi)}$ and ${Y_{\Oi} \in \mathfrak{X}(\Mi)}$ form a basis of $\mathfrak{T}$.\QEDbs	
\end{definition}

\begin{lemma}[Metric and curvature properties of PEMs]
	\label{lemma32}
	Suppose that	$u_{\iota} \in \mathcal{T}_{\omega_{\iota}} \mathcal{M}_{\iota}$ and $v_{\iota} \in \mathcal{T}_{\omega_{\iota}} \mathcal{M}_{\iota}$ are tangent vectors belonging to the tangent space $\mathcal{T}_{\omega_{\iota}} \mathcal{M}_{\iota}$ computed at ${{\omega_{\iota}} \in \mathcal{M}_{\iota}}$, $\forall \iota \in \mathcal{I}^m_{{G}_l}, m=1,2,\ldots,M$. Then, tangent vectors ${u_{G^m_l} \in \mathcal{T}_{\omega_{G^m_l}} \mathbb{M}_{G^m_l}}$ and ${v_{G^m_l} \in \mathcal{T}_{\omega_{G^m_l}} \mathbb{M}_{G^m_l}}$ are computed at $\omega_{G^m_l} \in \mathbb{M}_{G^m_l}$ by concatenation as ${u_{G^m_l} = (u_1, u_2, \cdots, u_{|\mathcal{I}^m_{{G}_l}|})}$ and 
	${v_{G^m_l} = (v_1, v_2, \cdots, v_{|\mathcal{I}^m_{{G}_l}|})}$. If each kernel submanifold $\mathcal{M}_{\iota}$ is endowed with a Riemannian metric $\mathfrak{d}_{\iota}$, then a $G^m_l$-PEM is endowed with the metric $\mathfrak{d}_{G^m_l}$ computed by
	\begin{equation}
	\mathfrak{d}_{G^m_l} ( u_{G^m_l} , v_{G^m_l} ) = \sum \limits _{\iota \in \mathcal{I}^m_{{G}_l}} \mathfrak{d}_{\iota}(u_{\iota},v_{\iota}).
	\label{eq:prod_metric}
	\end{equation}
	In addition, suppose that $\bar{C}_{\iota}$ is the Riemannian curvature tensor field (endomorphism) \cite{lee2009manifolds} of $\mathcal{M}_{\iota}$,  ${x_{\iota}, y_{\iota} \in \mathcal{T}_{\omega_{\iota}} \mathcal{M}_{\iota}}$, $\forall \iota \in \mathcal{I}^m_{{G}_l}$ defined by
	\begin{equation}
	\bar{C}_{\iota}(u_{\iota},v_{\iota},x_{\iota},y_{\iota}) = \left\langle {C}_{\iota} (U,V)X,Y \right\rangle_{\Oi}, 
	\label{eq:R_tensor}
	\end{equation}
	where $U,V,X,Y$ are vector fields such that $U_{\Oi} = u_{\iota} $, $V_{\Oi} = v_{\iota} $, $X_{\Oi} = x_{\iota} $, and $Y_{\Oi} = y_{\iota} $. Then, the Riemannian curvature tensor field $\bar{C}_{G_l} $ of $\mathbb{M}_{G_l}$ is computed by
	\begin{equation}
	\bar{C}_{G^m_l} ( u_{G^m_l} , v_{G^m_l}, x_{G^m_l} , y_{G^m_l}  ) = \sum \limits _{\iota \in \mathcal{I}^m_{{G}_l}} \bar{C}_{\iota}(u_{\iota},v_{\iota},x_{\iota},y_{\iota}),
	\label{eq:curv_tensor}
	\end{equation}
	where  ${x_{G^m_l} = (x_1, x_2, \cdots, x_{|\mathcal{I}^m_{{G}_l}|})}$ and 
	${y_{G^m_l} = (y_1, y_2, \cdots, y_{|\mathcal{I}^m_{{G}_l}|})}$. 
	Moreover, $\mathbb{M}_{G^m_l}$ has never strictly positive sectional curvature $\mathfrak{c}_{G^m_l}$ in the metric \eqref{eq:prod_metric}. In addition, if $\mathbb{M}_{G^m_l}$ is compact, then $\mathbb{M}_{G^m_l}$ does not admit a metric with negative sectional curvature $\mathfrak{c}_{G^m_l}$. \QEDbs
\end{lemma}


We compute the metric of a $G^m_l$-PEM $\mathbb{M}_{G^m_l}$ using the metrics identified on the component manifolds $\mathcal{M}_{\iota}$ employing \eqref{eq:prod_metric} given in Lemma~\ref{lemma32}. In addition, we use the Riemannian curvature and sectional curvature of the $\mathbb{M}_{G^m_l}$ given in Lemma~\ref{lemma32} to analyze convergence of our proposed G-SGD, and to compute adaptive step size. 

Note that some sectional curvatures vanish on the $\mathbb{M}_{G^m_l}$ by the lemma. For instance, suppose that each $\mathcal{M}_{\iota}$ is a unit two-sphere $\mathbb{S}^2$, $\forall \iota \in \mathcal{I}_{\mathcal{G}_l}$ (see Figure~\ref{fig1}.a). Then, $\mathbb{M}_{G^m_l}$ computed by \eqref{eq:prod_man} has unit curvature along two-dimensional subspaces of its tangent spaces, called two-planes. On the other hand, $\mathbb{M}_{G^m_l}$ has  zero curvature along all two-planes spanning exactly two distinct spheres. Therefore, learning rates need to be computed adaptively according to sectional curvatures at each layer of the CNN and at each epoch of the G-SGD for each kernel $\omega$ on each manifold $\mathbb{M}_{G^m_l}$. 

\subsection{Optimization using G-SGD in CNNs}

An algorithmic description of our proposed geometry-aware SGD (G-SGD) is given in Algorithm~\ref{alg1}. At the initialization of the G-SGD, we identify the component embedded kernel submanifolds $\mathcal{M}_{\iota}$ according to the constraints that will be applied on the kernels $\omega_{\iota} \in \mathcal{M}_{\iota}$. For instance, we employ an orthonormalization constraint $\| \omega_{\iota} \|_F = 1$ for kernels $\omega_{\iota}$ residing on $n_{\iota}$ dimensional unit sphere $\mathcal{M}_{\iota} \equiv \mathbb{S}^{n_{\iota}}$, where $\| \cdot \| _F$ is the Frobenius norm \cite{manopt_book}\footnote{In the experimental analyses, we use the oblique and the Stiefel manifolds as well as the sphere and the Euclidean space to identify subcomponent manifolds $\mathcal{M}_{\iota}$.}. 

When we employ a G-SGD on a $G^m_l$-PEM $\mathbb{M}_{G^m_l}$, each kernel $\omega_{G^m_l}^t \in \mathbb{M}_{G^m_l}$ is moved on the $G^m_l$-PEM in the descent direction of gradient of loss at each $t^{th}$ step of the G-SGD. More precisely, direction and amount of movement of a kernel $\omega^t_{G^m_l}$ are determined at the $t^{th}$ step and the $l^{th}$ layer by the following steps of Algorithm~\ref{alg1}:


\begin{algorithm}[tb]
	\caption{Optimization using G-SGD on an ensemble of PEMs.}
	\begin{algorithmic}[1]
		\STATE {\bfseries Input:} $T$ (number of iterations), $S$ (training set), \\ $\Theta$ (set of hyperparameters), $\mathcal{L}$ (a loss function), ${\mathcal{I}^m_{{G}_l} \subseteq \mathcal{I}_{\mathcal{G}_l}}, \forall m=1,2,\ldots,M, l=1,2,\ldots,L$.
		\STATE {\bfseries Initialization:} Construct an ensemble of products of kernel submanifolds~$ \mathcal{G}_l$, and initialize kernels
		${ \omega_{G^m_l}^t \in \mathbb{M}_{G^m_l} }$ using \eqref{eq:prod_man} with ${\mathcal{I}^m_{{G}_l} \subseteq \mathcal{I}_{\mathcal{G}_l}}, \forall m,l$.
		\FOR{each iteration $t=1,2,\ldots,T$}
		\FOR{each layer $l=1,2,\ldots,{L}$}
		
		\STATE Compute the gradient $\gr_E \mathcal{L}(\omega_{G^m_l}^{t}),\forall {G^m_l}$.
		
		
		\STATE ${
		\gr \mathcal{L}(\omega_{G^m_l}^{t}) := {\rm \Pi}_{\omega_l^t}  \Big ( \gr_E \; \mathcal{L}(\omega_{G^m_l}^{t}),\Theta \Big)},\forall {G^m_l}$.

		
		\STATE $ v_t := h(\gr \mathcal{L}(\omega_{G^m_l}^{t}), g(t,\Theta)), \forall {G^m_l}$.
		
		\STATE $
		\omega_{G^m_l}^{t+1} := \phi_{\omega_{G^m_l}^t}(  v_t), \forall \omega_{G^m_l}^t, \forall {G^m_l}$.

		\STATE $ t := t+1$.
		\ENDFOR
		\ENDFOR
		\STATE {\bfseries Output:} A set of estimated kernels $\{\omega_{G^m_l}^T \}_{l=1}^{{L}}, {\forall {G^m_l} \subseteq \mathcal{G}_l}$.
			\label{alg1}
			
	\end{algorithmic}	
	
\end{algorithm}



 
\begin{enumerate}[leftmargin=*]
	\item Line 6: Using Lemma~\ref{lemma32}, the gradient  $\gr_E \; \mathcal{L}(\omega_{G^m_l}^{t})$, which is obtained using back-propagation from the upper layer, is projected onto the tangent space $\mathcal{T}_{\omega^t_{G^m_l}} \mathbb{M}_{G^m_l}$ at $\gr \mathcal{L}(\omega_{G^m_l}^{t})$, where
	$\mathcal{T}_{\omega^t_{G^m_l}} \mathbb{M}_{G^m_l} = \bigtimes \limits _{\iota \in \mathcal{I}_{G^m_l}} \mathcal{T}_{\omega^t_{\iota,l}} \mathbb{M}_{\iota}$.

	\item Line 7: Movement of $\omega^t_{G^m_l}$ on $\mathcal{T}_{\omega^t_{G^m_l}} \mathbb{M}_{G^m_l}$ using $ h(\gr \mathcal{L}(\omega_{G^m_l}^{t}), g(t,\Theta))$ computed by
	\begin{equation}
	h(\gr \mathcal{L}(\omega_{G^m_l}^{t}), g(t,\Theta)) = -\frac{g(t,\Theta)}{\mathfrak{g}(\omega_{G^m_l}^t)}\gr \mathcal{L}(\omega_{G^m_l}^{t}),
	\label{eq:steps}
	\end{equation}
	where $g(t,\Theta)$ is the learning rate that satisfies
\begin{equation}
\sum_{t=0} ^{\infty} g(t,\Theta) = +\infty \; {\rm and} \; \sum_{t=0} ^{\infty} g(t,\Theta)^2 < \infty,
\label{eq:rate}
\end{equation} 
 $\mathfrak{g}(\omega_{G^m_l}^t) = \max\{ 1,\Gamma_1^t\}^{\frac{1}{2}}$, $\Gamma_1^t = (R_{G^m_l}^{t})^2 \Gamma_2^t$, ${\Gamma_2^t = \max \{(2\rho_{G^m_l}^{t} + R_{G^m_l}^{t})^2, (1+\mathfrak{c}_{G^m_l}(\rho_{G^m_l}^{t} + R_{G^m_l}^{t}))\} }$, $\rho_{G^m_l}^{t} \triangleq \rho(\omega_{G^m_l}^t, \hat{\omega}_{G^m_l}) $ is the geodesic distance between $\omega_{G^m_l}^t$ and a local minima $\hat{\omega}_{G^m_l}$ on $\mathbb{M}_{G^m_l}$, $\mathfrak{c}_{G^m_l}$ is the sectional curvature of $\mathbb{M}_{G^m_l}$, $R_{G^m_l}^{t} \triangleq  \| \gr \mathcal{L}(\omega_{G^m_l}^{t})  \|_2$ which can be computed using Lemma~\ref{lemma32} by
\begin{equation}
 \| \gr \mathcal{L}(\omega_{G^m_l}^{t})  \|_2 = \Big (\sum \limits_{\iota \in \mathcal{I}_{G^m_l}} \gr \mathcal{L}(\omega_{l,\iota}^{t})^2 \Big)^{\frac{1}{2}}.
 \label{eq:grad_norm}
\end{equation}

	\item Line 8: Projection of the moved kernel at $v_t$ onto the manifold $\mathbb{M}_{G^m_l}$ using $\phi_{\omega_{G^m_l}^t}(  v_t)$ to compute $\omega^{t+1}_{G^m_l}$, where $\phi_{\omega_{G^m_l}^t}(  v_t)$ is an exponential map, or a retraction which is an approximation of the exponential map \cite{absil_retr}.
\end{enumerate}   


The denominator  $\mathfrak{g}(\omega_{G^m_l}^t)$ used for computation of the step size in \eqref{eq:steps} is employed as a regularizer to control the change of gradient $\gr \mathcal{L}(\omega_{G^m_l}^{t})$ at each step of G-SGD. This property is examined in the experimental analyses for PEMs of different manifolds. For computation of $\mathfrak{g}(\omega_{G^m_l}^t)$, we use \eqref{eq:grad_norm} utilizing Lemma~\ref{lemma32}. Unlike related works, kernels residing on each PEM are moved and projected jointly on the PEMs in G-SGD, by which we can employ their interaction using the corresponding gradients considering nonlinear geometry of manifolds. G-SGD can perform optimization on PEMs and their ensemble according to sets $G^m_l, \forall m$, recursively. Thereby, G-SGD can consider interactions between component manifolds as well as those between PEMs in an ensemble. SGD methods studied in the literature do not have assurance of convergence when it is applied to optimization on ensembles of PEMs. Employment of \eqref{eq:steps} and \eqref{eq:rate} at line 7, and retractions at line 8 are essential for assurance of convergence as explained next. 
\subsection{Convergence Properties of G-SGD}
In some machine learning tasks, such as clustering \cite{sgdman,zhangSra16a}, the geodesic distance $\rho_{G^m_l}^{t}$ can be computed in closed form. However, a closed form solution may not be computed using CNNs due to the challenge of computation of local minima. Therefore, we provide an asymptotic convergence property for Algorithm~\ref{alg1} in the next theorem.

 \begin{theorem}
 	\label{thm33}
 	Suppose that there exists a local minimum $\hat{\omega}_{G^m_l} \in \mathbb{M}_{G^m_l}, \forall G^m_l \subseteq \mathcal{G}_l, \forall l$, and $\exists \epsilon>0$ such that $\inf \limits _{\rho_{G^m_l}^{t} > \epsilon^{\frac{1}{2}}} \left\langle \phi_{\omega_{G^m_l}^t}(\hat{\omega}_{G^m_l})^{-1}, \nabla \mathcal{L}(\omega_{G^m_l}^t) \right\rangle <0$, where $\phi$ is an exponential map or a twice continuously differentiable retraction, and $\langle \cdot,\cdot \rangle$ is the inner product. The loss function and the gradient converges almost surely (a.s.) by $\mathcal{L}(\omega^t_{G^m_l}) \xrightarrow[t \to \infty]{\rm a.s.} \mathcal{L}(\hat{\omega}_{G^m_l})$, and $\nabla \mathcal{L}(\omega^t_{G^m_l}) \xrightarrow[t \to \infty]{\rm a.s.} 0$, for each $\mathbb{M}_{G^m_l}, \forall l$. \QEDbs
 	
  \end{theorem}

Theorem~\ref{thm33} assures convergence of the G-SGD (Algorithm~\ref{alg1}) to minima. For  implementation of G-SGD, we use the result given in Lemma~\ref{lemma32} for PEMs to employ sectional curvatures. Although sectional curvatures of non-identical embedded kernel submanifolds can be different \cite{oo16}, Lemma~\ref{lemma32} assures existence of zero sectional curvature in PEMs along their tangent spaces. In the next theorem, we provide an example for computation of a step size function $\mathfrak{g}(\cdot)$ for component embedded kernel submanifolds determined by the sphere using the result given in Lemma~\ref{lemma32}, and explore its convergence property using Theorem~\ref{thm33}.

\begin{corr}
	\label{corr34}
Suppose that $\mathbb{M}_{\iota}$ are identified by ${n_{\iota} \geq 2}$ dimensional unit sphere $\mathbb{S}^{n_{\iota}}$, and $\rho_{G^m_l}^t \leq \hat{\mathfrak{c}}^{-1}$, where $\hat{\mathfrak{c}}$ is an upper bound on the sectional curvatures of $\mathbb{M}_{G^m_l}, \forall l$ at $\omega_{G^m_l}^t \in \mathbb{M}_{G^m_l}, \forall t$. If step size is computed using \eqref{eq:steps} with 
\begin{equation}
{\mathfrak{g}(\omega_{G^m_l}^t) = (\max\{ 1, (R_{G^m_l}^{t})^2(2+R_{G^m_l}^{t})^2 \} })^{\frac{1}{2}},
\end{equation}
 then ${\mathcal{L}(\omega^t_{G^m_l}) \xrightarrow[t \to \infty]{\rm a.s.} \mathcal{L}(\hat{\omega}_{G^m_l})}$, and ${\nabla \mathcal{L}(\omega^t_{G^m_l}) \xrightarrow[t \to \infty]{\rm a.s.} 0}$, for each $\mathbb{M}_{G^m_l}, \forall l$. \QEDbs	
\end{corr}

In the experimental analyses, we use different step size functions and analyze convergence properties and performance of CNNs trained using G-SGD by relaxing assumptions of  Theorem~\ref{thm33} and Corollary~\ref{corr34} for different CNN architectures and benchmark image classification datasets.

\section{Experimental Analyses}
\label{sec:exp}
 
We examine the proposed G-SGD method for training of state-of-the-art CNNs, called Residual Networks (Resnets) \cite{res_net}, equipped with different number of layers and kernels. We use three benchmark RGB image classification datasets, namely Cifar-10, Cifar-100 and Imagenet \cite{Alexnet}. The Cifar-10 and Cifar-100 datasets consist of $5\times 10^4$  training and $10^4$ test images belonging to 10 and 100 classes, respectively. The Imagenet dataset consists   of   $10^3$ classes  ($12 \times 10^4$ training and $5\times 10^4$ validation images). 

We construct ensembles of PEMs using the sphere (Sp), the oblique (Ob) and the Stiefel (St) manifolds. We also use the kernels residing on the ambient Euclidean space of embedded kernel submanifolds (Euc.). In order to preserve the task structure (classification of RGB images), we employed PI for the layers $l=2,3,\ldots,L$ considering the RGB space of images, PO for  $l=1,2,\ldots,L-1$ considering the number of classes learned at the top $L^{th}$ layer of a CNN, and PIO for $l=2,\ldots,L-1$. Suppose that we have a set of $N_l$ kernels $\mathfrak{N}_l$ with $|\mathfrak{N}_l| = N_l$ and $|\mathcal{I}_{\mathcal{G}_l}| = N_l$ at the $l^{th}$ layer of a CNN. In the construction of ensembles, we employ PI, PO and PIO using a kernel set splitting (KSS) scheme. In KSS, we split the kernel set $\mathfrak{N}_l$ into $M$ subsets ${\mathfrak{N}_l^m \subset \mathfrak{N}_l}$, $ \forall m=1,2,\ldots,M$, where kernels ${\omega \in \mathfrak{N}^m_l}$ belonging to $\mathfrak{N}_l^m$ reside on the $m^{th}$ PEM $\mathbb{M}_{G^m_l}$ identified by $\mathcal{I}_{G^m_l}$ which is determined according to PI, PO and PIO, $\forall m$. For the sake of simplicity of the analyses, we split the kernel set into  subsets with size $\frac{N_l}{M}$ in KSS, while the proposed schemes enable us to construct new kernel sets with varying size. Implementation details of G-SGD for different ensembles and Resnets, data pre-processing details of the benchmark datasets and additional results are given in the supp. mat.

\begin{table}[t]
	\centering
	\caption{Results for Resnet-44 on Cifar-10 with DA.}
	\begin{tabular}{C{4.95cm}C{2.5cm}}
		\toprule
		\toprule
		
		\textbf{Model} & \textbf{Class. Error(\%)} \\
		\midrule
		\midrule
		
		Euc. \cite{res_net}  & 7.17  \\
		Euc. \cite{oo16} & 7.16  \\
		Euc. $\dagger$ & {\color{red} 7.05}  \\
		Sp/Ob/St \cite{oo16} & 6.99/6.89/{{6.81}}\\
		Sp/Ob/St  $\dagger$ & 6.84/6.87/{ {6.73}}\\
		PEMs of	Sp/Ob/St   & 6.81/6.85/{ {6.70}}\\
	PI for PEMs of	Sp/Ob/St   & 6.82/6.81/{ {6.70}}\\
	  PI (Euc.+Sp/Euc.+St/Euc.+Ob) & 6.89/6.84/6.88  \\
		PI (Sp+Ob/Sp+St/Ob+St) & 6.75/6.67/6.59 \\	
		PI (Sp+Ob+St/Sp+Ob+St+Euc.)  & 6.31/6.34  \\	
		PO for PEMs of	Sp/Ob/St  & 6.77/6.83/{ {6.65}}\\	
		PO (Euc.+Sp/Euc.+St/Euc.+Ob) & 6.85/6.78/6.90  \\		
		PO (Sp+Ob/Sp+St/Ob+St) & 6.62/6.59/6.51  \\	
		PO (Sp+Ob+St/Sp+Ob+St+Euc.)  & 6.35/6.22 \\		
	PIO for PEMs of	Sp/Ob/St   & 6.71/6.73/{ {6.61}}\\					
		PIO (Euc.+Sp/Euc.+St/Euc.+Ob) & 6.95/6.77/6.82  \\		
PIO (Sp+Ob/Sp+St/Ob+St) & 6.21/6.19/6.25  \\
PIO (Sp+Ob+St/Sp+Ob+St+Euc.)  & 5.95/{\color{blue} 5.92 } \\		
		\bottomrule
		\bottomrule
	\end{tabular}%
	\label{tab:res10}%
\end{table}%

\begin{table}[t]
	\centering
	\caption{Results for Resnet-18 which are trained using Imagenet for single crop validation error rate (\%).}		
	\begin{tabular}{C{4.85cm} C{2.70cm}}
		\toprule
		\toprule
		
		\textbf{Model} & \textbf{Top-1 Error (\%)} \\
		
		\midrule
		\midrule
		Euc. \cite{oo16} & 30.59\\
		Euc. $\dagger$ & {\color{red} 30.31}\\
		Sp/Ob/St\cite{oo16}  & 29.13/28.97/{{28.14}}\\
		Sp/Ob/St $\dagger$  & 28.71/28.83/{{28.02}}\\	
		PEMs of Sp/Ob/St & 28.70/28.77/{{28.00}}\\	
PI for PEMs of Sp/Ob/St & 28.69/28.75/{{27.91}}\\				
		PI (Euc.+Sp/Euc.+St/Euc.+Ob) & 30.05/29.81/29.88  \\		
		PI (Sp+Ob/Sp+St/Ob+St) & 28.61/28.64/28.49  \\
		PI (Sp+Ob+St/Sp+Ob+St+Euc.)  & 27.63/27.45  \\	
PO for PEMs of Sp/Ob/St & 28.67/28.81/{{27.86}}\\						
		PO (Euc.+Sp/Euc.+St/Euc.+Ob) & 29.58/29.51/29.90  \\		
		PO (Sp+Ob/Sp+St/Ob+St) & 28.23/28.01/28.17  \\
		PO (Sp+Ob+St/Sp+Ob+St+Euc.)  & 27.81/27.51  \\
PIO for PEMs of Sp/Ob/St & 28.64/28.72/{{27.83}}\\								
		PIO (Euc.+Sp/Euc.+St/Euc.+Ob) & 29.19/28.25/28.53  \\		
PIO (Sp+Ob/Sp+St/Ob+St) & 28.14/27.66/27.90  \\
PIO (Sp+Ob+St/Sp+Ob+St+Euc.)  & 27.11/{\color{blue} 27.07}  \\			
		\bottomrule
		\bottomrule
	\end{tabular}%
	\label{tab:imagenet}%
\end{table}%

\begin{table*}[ht]
	\centering
	\caption{Classification error (\%) for training 110-layer Resnets with constant depth (RCD) and Resnets with stochastic depth (RSD) using the PIO scheme on Cifar-10 and Cifar-100, with and without using DA.}	
	\begin{tabular}{C{4.8cm}|C{1.7cm}|C{2.250cm}|C{2.69cm}|C{2.69cm}|}
		\toprule
		\toprule
		\multicolumn{1}{c}{\textbf{Model}} & \multicolumn{1}{c}{\textbf{Cifar-10 w. DA}} & \multicolumn{1}{c}{\textbf{Cifar-100 w. DA}}  & \multicolumn{1}{c}{\textbf{Cifar-10 w/o DA}} & \multicolumn{1}{c}{\textbf{Cifar-100 w/o DA}} \\
		\bottomrule
		RCD \cite{DCCN}  & 6.41 & 27.22 & 13.63  & 44.74 \\
		(Euc.) $\dagger$  & {\color{red} 6.30} &{\color{red} 27.01} &{\color{red} 13.57}  & {\color{red} 44.65 }\\		
		Sp/Ob/St (\cite{oo16}) & 6.22/6.07/{{5.93}} & 26.44/25.99/{{25.41}} & 13.11/12.94/{{12.88}} & 42.51/42.30/{{40.11}} \\
		Sp/Ob/St $\dagger$  & 6.05/6.03/{{5.91}} & 26.19/25.87/{{25.39}} & 12.96/12.85/{{12.79}} & 42.13/42.00/{{39.94}} \\
		PEMs of Sp/Ob/St & 6.00/6.01/{{5.86}} & 25.93/25.74/{{25.18}} & 12.74/12.77/{{12.74}} & 42.02/42.88/{{39.90}} \\		
		PIO for PEMs of Sp/Ob/St  & 5.95/5.91/{{5.83}} & 25.89/25.71/{{25.12}} & 12.71/12.72/{{12.69}} & 41.68/42.75/{{39.83}} \\
		PIO (Euc.+Sp/Euc.+St/Euc.+Ob)  & 6.03/5.99/6.01  & 25.57/25.49/25.64  & 12.77/12.21/12.92  & 41.90/41.37/41.85 \\	
		PIO   (Sp+Ob/Sp+St/Ob+St)  & 5.97/5.86/5.46 & 24.71/24.96/24.76  & 11.47/11.65/ 11.51  & 41.49/40.53/40.34  \\		
		PIO (Sp+Ob+St/Sp+Ob+St+Euc.)  & 5.25/{\color{blue} 5.17}  & 23.96/{\color{blue} 23.79 }   & 11.29/{\color{blue} 11.15}  & 39.53/ {\color{blue} 39.35 } \\
		\bottomrule
		RSD \cite{DCCN}   & 5.23  & 24.58 & 11.66 & 37.80  \\
		Euc. $\dagger$   & {\color{red} 5.17 } & {\color{red} 24.39} & {\color{red} 11.40} & {\color{red} 37.55 } \\		
		Sp/Ob/St \cite{oo16} & 5.20/5.14/{{4.79}}  & 23.77/23.81/{{23.16}} & 10.91/10.93/{{10.46}} & 36.90/36.47/{{35.92}} \\
		Sp/Ob/St $\dagger$ & 5.08/5.11/{{4.73}}  & 23.69/23.75/{{23.09}} & 10.52/10.66/{{10.33}} & 36.71/36.38/{{35.85}} \\
		PEMs of Sp/Ob/St & 5.05/5.08/{{4.69}}  & 23.51/23.60/{{23.85}} & 10.41/10.54/{{10.25}} & 36.40/36.11/{{35.53}} \\
		PIO for PEMs of Sp/Ob/St & 4.95/5.03/{{4.62}}  & 23.47/23.51/{{23.77}} & 10.37/10.51/{{10.19}} & 36.33/36.02/{{35.41}} \\		
		PIO   (Euc.+Sp/Euc.+St/Euc.+Ob)  & 5.00/5.08/5.14  & 23.69/23.25/23.32  & 10.74/10.25/10.93  & 35.76/35.55/35.81 \\	
		PIO   (Sp+Ob/Sp+St/Ob+St)  & 4.70/4.58/4.90  & 22.84/22.91/22.80  & 10.13/10.24/10.06  & 35.66/35.01/35.35 \\		
		PIO (Sp+Ob+St/Sp+Ob+St+Euc.)  & {\color{blue} 4.29}/4.31   & 22.19/{\color{blue} 22.03}  &  {\color{blue} 9.52}/9.56 & 34.49/{\color{blue} 34.25} \\
		\bottomrule
		\bottomrule
								
	\end{tabular}%
	\label{tab:rcd}%
\end{table*}%

\subsection{Analysis of Classification Performance on Benchmark Datasets}

We analyze classification performance of CNNs trained using G-SGD on benchmark Cifar-10, Cifar-100 and Imagenet datasets. In order to construct ensembles of kernels belonging to Euc., Sp, St and Ob using KSS, we increase the number of kernels used in CNNs to 24 and its multiples (see the supp. mat.). We use other hyperparameters of CNNs as suggested in  \cite{res_net,SN,oo16}. We depict performance of our implementation of  CNNs for baseline geometries (Euc., Sp, St and Ob) by $\dagger$ marker in the tables. For computation of  $\mathfrak{g}(\omega_{G^m_l}^t)$, we used 
\begin{equation}
{\mathfrak{g}(\omega_{G^m_l}^t) = (\max\{ 1, (R_{G^m_l}^{t})^2(2+R_{G^m_l}^{t})^2 \} })^{\frac{1}{2}}, \forall m,l
\end{equation}
as suggested in Corollary~\ref{corr34}. Implementation details are given in the supp. mat.  

We examine classification performance of Resnets with 44 layers (Resnet-44) and 18 layers (Resnet-18) on Cifar-10  with data augmentation (DA) and Imagenet in Table~\ref{tab:res10} and Table~\ref{tab:imagenet}, respectively. The results show that performance of CNNs are boosted by employing ensembles of PEMs (denoted by PI, PO and PIO for PEMs) using G-SGD compared to the employment of baseline Euc. We observe that PEMs of component submanifolds of identical geometry (denoted by PEMs of Sp/St/Ob), and their ensembles (denoted by PI, PO, PIO for PEMs of Sp/St/Ob)  provide better performance compared to employment of component submanifolds (denoted by Sp/Ob/St) \cite{oo16}. For instance, we obtain $28.64\%$, $28.72\%$ and $27.83\%$ error using PIO for PEMs of Sp, Ob and St in Table~\ref{tab:imagenet}, respectively. However, the error obtained using Sp, Ob and St is $28.71\%$, $28.83\%$ and $28.02\%$, respectively.

In addition, we obtain $0.28\%$ and $2.06\%$ boost of the performance by ensemble of the St with Euc. ($6.77\%$ and $28.25\%$ using PIO for Euc.+St, respectively) for the experiments on the Cifar-10 and Imagenet datasets using the  PIO scheme in Table~\ref{tab:res10} and Table~\ref{tab:imagenet}, respectively. Moreover, we observe that construction of ensembles using Ob performs better for PI compared to PO. For instance, we observe that PI for PEMs of Ob provides $6.81\%$ and $28.75\%$ while PO for PEMS of Ob provides $6.83\%$ and $28.81\%$ in Table~\ref{tab:res10} and Table~\ref{tab:imagenet}, respectively.   We may associate this result with the observation that kernels belonging to Ob are used for feature selection and modeling of texture patterns with high performance \cite{oblq,oo16}. However, ensembles of St and Sp perform better for PO ($6.59\%$ and $28.01\%$ in Table~\ref{tab:res10} and Table~\ref{tab:imagenet}) compared to PI ($6.67\%$ and $28.64\%$ in Table~\ref{tab:res10} and Table~\ref{tab:imagenet}) on kernels employed on output channels. 

It is also observed that  PIO performs better than PI and PO in all the experiments. We observe {\color{blue} $1.13\%$} and {\color{blue} $3.24\%$} boost by construction of an ensemble of four manifolds (Sp+Ob+St+Euc.) using the  PIO scheme in Table~\ref{tab:res10}  ({\color{blue} $5.92\%$}) and Table~\ref{tab:imagenet} ({\color{blue} $27.07\%$}), respectively. In other words, ensemble methods boost the performance of large-scale CNNs more for large-scale datasets (e.g. Imagenet) consisting of larger number of samples and classes compared to the performance of smaller CNNs employed on smaller datasets (e.g. Cifar-10). This result can be attributed to enhancement of sets of features learned using multiple constraints on kernels.

We analyze this observation by examining the performance of larger CNNs consisting of 110 layers on Cifar-10 and Cifar-100 datasets with and without using DA in Table~\ref{tab:rcd}. The results show that employment of PEMs can boost the performance of CNNs that use component submanifolds (e.g. PEMs of Sp, Ob and St) more for larger networks (Table~\ref{tab:rcd}) compared to smaller networks (Table~\ref{tab:res10} and Table~\ref{tab:imagenet}).  Moreover, employment of PIO for PEMs of Sp+Ob+St+Euc. boosts the performance of CNNs that use {\color{red}Euc.} more for Cifar-100 (3.55\% boost in average) compared to the performance obtained for Cifar-10 (1.58\% boost in average). In addition, we observe that ensembles boost the performance of CNNs that use DA methods more compared to the performance of CNNs without using DA. 

Our method fundamentally differs from network ensembles. In order to analyze the results for network ensembles of CNNs, we employed an ensemble method \cite{res_net} by \textit{voting of decisions} of Resnet 44 on Cifar 10. When CNNs trained on individual Euc, Sp, Ob, and St are ensembled using voting, we obtained $7.02\%$  (Euc+Sp+Ob+St) and $6.85\%$ (Sp+Ob+St) errors (see Table 1 for comparison). In our analyses of ensembles (PI, PO and PIO), each PEM contains $\frac{N_l}{M}$ kernels, where $N_l$ is the number of kernels used at the $l^{th}$ layer, and $M$ is the number of PEMs. When each CNN in the ensemble was trained using an individual manifold which contains $\frac{1}{4}$ of kernels (using $M=4$ as utilized in our experiments), then we obtained $11.02\%$ (Euc), $7.76\%$ (Sp), $7.30\%$ (Ob), $7.18\%$ (St), $9.44\%$ (Euc+Sp+Ob+St) and $7.05\%$ (Sp+Ob+St) errors. Thus, our proposed methods outperform ensembles constructed by voting. Additional results are given in the supplemental material.

\section{Conclusion and Discussion}
\label{sec:conc}

We introduced and elucidated a problem of training CNNs using multiple constraints  employed on  convolution kernels with convergence properties. Following our theoretical results, we proposed the G-SGD algorithm and adaptive step size estimation methods for optimization on ensembles of PEMs that are identified by the constraints. The experimental results show that our proposed  methods can improve   convergence properties and classification performance of CNNs. Overall, the results show that employment of ensembles of PEMs using G-SGD can boost the performance of larger CNNs (e.g. RCD and RSD) on large scale datasets (e.g. Imagenet) more compared to the performance of small and medium scale networks (e.g. Resnets with 16 and 44 layers) employed on smaller datasets (e.g. Cifar-10). 

In future work, we plan to extend the proposed framework by development of new ensemble schemes to perform various tasks such as machine translation and video recognition using CNNs and Recurrent Neural Networks (RNNs). In addition, the proposed methods can be applied to other stochastic optimization methods such as Adam and trust region methods. We believe that our proposed framework will be useful for researchers to study geometric properties of parameter spaces of deep networks, and to improve our understanding of deep feature representations.

\bibliography{myref}

\begin{thebibliography}{10}\itemsep=-1pt

\bibitem{oblq}
P.~A. Absil and K.~A. Gallivan.
\newblock Joint diagonalization on the oblique manifold for independent
  component analysis.
\newblock In {\em Proc. 31st IEEE Int. Conf. Acoust., Speech Signal Process},
  volume~5, pages 945--948, Toulouse, France, May 2006.

\bibitem{manopt_book}
P.-A. Absil, R.~Mahony, and R.~Sepulchre.
\newblock {\em Optimization Algorithms on Matrix Manifolds}.
\newblock PUP, Princeton, NJ, USA, 2007.

\bibitem{absil_retr}
P.~A. Absil and J.~Malick.
\newblock Projection-like retractions on matrix manifolds.
\newblock {\em SIAM Journal on Optimization}, 22(1):135--158, 2012.

\bibitem{unit_evol}
M.~Arjovsky, A.~Shah, and Y.~Bengio.
\newblock Unitary evolution recurrent neural networks.
\newblock In {\em Proc. of the 33rd Int. Conf. on Mach. Learn. {ICML}}, pages
  1120--1128, New York City, NY, USA, June 2016.

\bibitem{norm_prop}
D.~Arpit, Y.~Zhou, B.~U. Kota, and V.~Govindaraju.
\newblock Normalization propagation: {A} parametric technique for removing
  internal covariate shift in deep networks.
\newblock In {\em Proc. of the 33rd Int. Conf. on Mach. Learn. {ICML}}, pages
  1168--1176, New York City, NY, USA, June 2016.

\bibitem{sgdman}
S.~Bonnabel.
\newblock Stochastic gradient descent on riemannian manifolds.
\newblock {\em IEEE Trans. Autom. Control}, 58(9):2217--2229, Sept 2013.

\bibitem{RBN}
M.~Cho and J.~Lee.
\newblock Riemannian approach to batch normalization.
\newblock In {\em Advances in Neural Information Processing Systems (NIPS)},
  2017.

\bibitem{parseval}
M.~Cisse, P.~Bojanowski, E.~Grave, Y.~Dauphin, and N.~Usunier.
\newblock Parseval networks: Improving robustness to adversarial examples.
\newblock In D.~Precup and Y.~W. Teh, editors, {\em Proceedings of the 34th
  International Conference on Machine Learning}, volume~70 of {\em Proceedings
  of Machine Learning Research}, pages 854--863, International Convention
  Centre, Sydney, Australia, 06--11 Aug 2017. PMLR.

\bibitem{res_net}
K.~He, X.~Zhang, S.~Ren, and J.~Sun.
\newblock Deep residual learning for image recognition.
\newblock In {\em IEEE Int. Conf. on Comp. Vis. Patt. Recog. {(CVPR)}}, 2016.

\bibitem{Henaff}
M.~Henaff, A.~Szlam, and Y.~LeCun.
\newblock Recurrent orthogonal networks and long-memory tasks.
\newblock In {\em Proc. of the 33rd Int. Conf. Mach. Learn.}, ICML'16, pages
  2034--2042, 2016.

\bibitem{DCCN}
G.~Huang, Z.~Liu, and K.~Q. Weinberger.
\newblock Densely connected convolutional networks.
\newblock In {\em Proc. IEEE Conf. Comp. Vis. Patt. Recog. (CVPR)}, 2017.

\bibitem{SN}
G.~Huang, Y.~Sun, Z.~Liu, D.~Sedra, and K.~Q. Weinberger.
\newblock Deep networks with stochastic depth.
\newblock In {\em Proc. of the 14th European Conf. Comp. Vis.}, pages 646--661,
  Amsterdam, The Netherlands, 2016.

\bibitem{Huang_2017_ICCV}
L.~Huang, X.~Liu, Y.~Liu, B.~Lang, and D.~Tao.
\newblock Centered weight normalization in accelerating training of deep neural
  networks.
\newblock In {\em The IEEE International Conference on Computer Vision (ICCV)},
  Oct 2017.

\bibitem{haaai}
Z.~Huang and L.~V. Gool.
\newblock A riemannian network for spd matrix learning.
\newblock In {\em Assoc. for the Adv. of Artificial Intelligence (AAAI)}, Feb
  2017.

\bibitem{huang162}
Z.~Huang, C.~Wan, T.~Probst, and L.~V. Gool.
\newblock Deep learning on lie groups for skeleton-based action recognition.
\newblock In {\em Proc. IEEE Conf. Comp. Vis. Patt. Recog. (CVPR)}, 2017.

\bibitem{icml2015_huanga15}
Z.~Huang, R.~Wang, S.~Shan, X.~Li, and X.~Chen.
\newblock Log-euclidean metric learning on symmetric positive definite manifold
  with application to image set classification.
\newblock In {\em Proc. of the 32nd Int. Conf. Mach. Learn. (ICML-15)}, pages
  720--729, 2015.

\bibitem{SNN}
G.~Klambauer, T.~Unterthiner, A.~Mayr, and S.~Hochreiter.
\newblock Self-normalizing neural networks.
\newblock In {\em Advances in Neural Information Processing Systems (NIPS)},
  2017.

\bibitem{Alexnet}
A.~Krizhevsky, I.~Sutskever, and G.~E. Hinton.
\newblock Imagenet classification with deep convolutional neural networks.
\newblock In {\em Advances in Neural Information Processing Systems (NIPS)},
  pages 1097--1105, 2012.

\bibitem{IGAN}
A.~Kumar, P.~Sattigeri, and T.~Fletcher.
\newblock Improved semi-supervised learning with gans using manifold
  invariances.
\newblock In {\em Advances in Neural Information Processing Systems (NIPS)},
  2017.

\bibitem{lee2009manifolds}
J.~Lee.
\newblock {\em Manifolds and Differential Geometry}.
\newblock Graduate studies in mathematics. American Mathematical Society, 2009.

\bibitem{oo16}
M.~Ozay and T.~Okatani.
\newblock Optimization on submanifolds of convolution kernels in cnns.
\newblock {\em CoRR}, abs/1610.07008, 2016.

\bibitem{w_norm}
T.~Salimans and D.~P. Kingma.
\newblock Weight normalization: A simple reparameterization to accelerate
  training of deep neural networks.
\newblock In {\em Advances in Neural Information Processing Systems (NIPS)},
  2016.

\bibitem{AAAI1714830}
W.~Shang, J.~Chiu, and K.~Sohn.
\newblock Exploring normalization in deep residual networks with concatenated
  rectified linear units.
\newblock In {\em AAAI Conference on Artificial Intelligence}, 2017.

\bibitem{zhangSra16a}
H.~Zhang and S.~Sra.
\newblock First-order methods for geodesically convex optimization.
\newblock In {\em Conference on Learning Theory (COLT)}, Jun. 2016.

\end{thebibliography}
\bibliographystyle{ieee}

\end{document}